\documentclass{article} 
\usepackage{iclr2021_conference,times}

\usepackage{amsmath,amsfonts,bm}

\def\eqref#1{equation~\ref{#1}}

\def\1{\bm{1}}

\DeclareMathAlphabet{\mathsfit}{\encodingdefault}{\sfdefault}{m}{sl}
\SetMathAlphabet{\mathsfit}{bold}{\encodingdefault}{\sfdefault}{bx}{n}

\usepackage{hyperref}
\usepackage{url}
\usepackage{graphicx}
\usepackage{algorithmicx,algorithm}
\usepackage{algpseudocode}
\usepackage{wrapfig}
\usepackage{booktabs}
\usepackage{caption}

\def \norm#1{\Vert #1 \Vert_2}
\def \bs#1{\boldsymbol{#1}}
\def \trans#1{#1^{\mathsf{T}}}

\newcommand\Algphase[1]{%
\vspace*{-.7\baselineskip}\Statex\hspace*{\dimexpr-\algorithmicindent-2pt\relax}\rule{\textwidth}{0.4pt}%
\Statex\hspace*{-\algorithmicindent}\textbf{#1}%
\vspace*{-.7\baselineskip}\Statex\hspace*{\dimexpr-\algorithmicindent-2pt\relax}\rule{\textwidth}{0.4pt}%
}

\title{FixNorm: Dissecting Weight Decay for Training Deep Neural Networks}

\author{Yucong Zhou, Yunxiao Sun \& Zhao Zhong \\
Huawei \\
\texttt{\{zhouyucong1, sunyunxiao3, zorro.zhongzhao\}@huawei.com} \\
}

\iclrfinalcopy 
\begin{document}

\maketitle

\begin{abstract}
Weight decay is a widely used technique for training Deep Neural Networks(DNN).
It greatly affects generalization performance but the underlying mechanisms are not fully understood.
Recent works show that for layers followed by normalizations, weight decay mainly affects the \emph{effective learning rate}.
However, despite normalizations have been extensively adopted in modern DNNs, layers such as the final fully-connected layer do not satisfy this precondition.
For these layers, the effects of weight decay are still unclear.
In this paper, we comprehensively investigate the mechanisms of weight decay and find that except for influencing effective learning rate, weight decay has another distinct mechanism that is equally important: affecting generalization performance by controlling \emph{cross-boundary risk}.
These two mechanisms together give a more comprehensive explanation for the effects of weight decay.
Based on this discovery, we propose a new training method called \textbf{FixNorm}, which discards weight decay and directly controls the two mechanisms.
We also propose a simple yet effective method to tune hyperparameters of FixNorm, which can find near-optimal solutions in a few trials.
On ImageNet classification task, training EfficientNet-B0 with FixNorm achieves 77.7\%, which outperforms the original baseline by a clear margin.
Surprisingly, when scaling MobileNetV2 to the same FLOPS and applying the same tricks with EfficientNet-B0, training with FixNorm achieves 77.4\%, which is only 0.3\% lower.
A series of SOTA results show the importance of well-tuned training procedures, and further verify the effectiveness of our approach.
We set up more well-tuned baselines using FixNorm, to facilitate fair comparisons in the community.
\end{abstract}

\section{Introduction}

Weight decay is an important trick and has been widely used in training Deep Neural Networks.
By constraining the weight magnitude, it is widely believed that weight decay regularizes the model and
improve generalization performance\citep{krizhevsky2012imagenet,bos1996optimal,bos1996using,krogh1992simple}.
Recently, a series of works \cite{van2017l2, zhang2018three, hoffer2018norm} propose that for layers followed by normalizations,
such as BatchNormalization\citep{ioffe2015batch}, 
the main effect of weight decay is increasing the \emph{effective learning rate}(ELR).
Take the widely adopted Conv-BN block as an example.
Since BatchNormalization is scale-invariant, the weight norm of the convolution layer do not actually affect the block's output,
which contradicts the regularization effect of weight decay.
On the contrary, the weight norm affects the step size of weight direction, e.g., larger weight norm results in smaller step size.
By constraining the weight norm from unlimitedly growing, weight decay is actually increasing the step size of weight direction,
thus increasing the ELR.

This interesting discovery arouses our questions: Does it covers the full mechanism(s) of weight decay?
If the answer is true, weight decay would be unnecessary since its effect fully overlaps with the original learning rate.
\citet{hoffer2018norm} shows that the performance can be recovered when training without weight decay by applying an LR correction technique.
However, the LR correction coefficient at each step depends on the original training statistics(with weight decay), which makes this technique only a verification of the effective learning rate hypothesis, but can not be used as a practical training method.
Moreover, the ELR hypothesis only applies to layers followed by normalizations, and there are layers that do not satisfy this requirement.
For example, the final fully-connected layer that commonly used in classification tasks.
For these layers, the effects of weight decay are usually omitted \citep{hoffer2018norm}, or simply the original weight decay is preserved \citep{zhang2018three}.
These problems indicate that the mechanisms of weight decay are still not fully understood.

In this paper we try to investigate the above problems.
For the convolution layers that followed by normalizations, we find that simply fixing the overall weight norm
to a constant fully recovers the effect of weight decay.
For the final fully-connected layer, we find that there is a special effect introduced by weight decay,
which influences the generalization performance by controlling the \textbf{cross-boundary risk}.
This mechanism is as important as the former investigated ELR, and they together capture most of the effects of weight decay.
These two mechanisms are unified into a new training scheme called \textbf{FixNorm},
which discards the weight decay and directly controls the effects of two main mechanisms.
By using FixNorm, we fully recover the performance of popular CNNs on large scale classification dataset ImageNet\citep{deng2009imagenet}.
Further, we show that the hyperparameters of FixNorm can be easily tuned, and propose a simple yet effective tuning method which
only requires a few trials and achieves near-optimal performance.
Specifically, by applying tuned FixNorm, we achieve 77.7\%(+0.4\%) with EfficientNet-B0, 79.5\%(+0.3\%) with EfficientNet-B1, 73.97\%(+1.9\%) with MobileNetV2.

Training tricks and network tricks show great impacts on performance, which also introduce difficulties in tuning and bring ambiguities to comparisons.
We show that this can be mitigated by using FixNorm.
For example, by simply scaling MobileNetV2 to the same FLOPS and applying the same tricks of EfficientNet-B0,
training with FixNorm achieves 77.4\% top-1 accuracy, while the default training process only get 76.72\%.
To facilitate fairer comparisons, we apply our FixNorm method to representative CNN architectures and set up new baselines under different settings.

Our contributions can be summarized as follows:
\begin{itemize}
  \item Except for increasing the effective learning rate, we discover a new mechanism of weight decay which controls the cross-boundary risk,
  and give a better understanding of weight decay's effect on generalization performance
  \item We propose a new training scheme called FixNorm that discards the weight decay and directly controls the effects of two main mechanisms,
  which not only fully recovers the accuracy of weight decay training, but also makes hyperparameters easier to tune.
  \item We propose a simple yet effective method to tune hyperparameters of FixNorm and demonstrate efficiency, robustness and SOTA performance on 
  large scale datasets like ImageNet, MS COCO\citep{lin2014microsoft} and Cityscapes\citep{Cordts2016Cityscapes}
  \item By using our approach, we establish well-tuned baselines for popular networks, which we hope can facilitate fairer comparisons in the community
\end{itemize}

\section{Dissecting weight decay for training Deep Neural Networks} \label{sec_main}

\subsection{Revisiting the effective learning rate hypothesis} \label{sec_revisit}
We aim at further understanding the mechanisms of weight decay. Towards this, we first briefly revisit the effective learning rate(ELR) hypothesis.
As noted in \citet{hoffer2018norm}, when BN is applied after a linear layer, the output is invariant to the channel weight vector norm.
Denoting a channel weight vector with $\boldsymbol{w}$ and $\hat{\boldsymbol{w}}=\boldsymbol{w} / \norm{\boldsymbol{w}}$,
channel input as $\boldsymbol{x}$, we have
\begin{equation}
  \text{BN}(\norm{\boldsymbol{w}} \hat{\boldsymbol{w}} \boldsymbol{x}) = \text{BN}(\hat{\boldsymbol{w}} \boldsymbol{x})
\end{equation}
In such case, the gradient is scaled by $1 / \norm{\boldsymbol{w}}$:
\begin{equation}
  \frac{\partial \text{BN}(\norm{\boldsymbol{w}} \hat{\boldsymbol{w}} \boldsymbol{x})}{\partial(\norm{\boldsymbol{w}} \hat{\boldsymbol{w}})} = \frac{1}{\norm{\boldsymbol{w}}} \frac{\partial \text{BN}(\hat{\boldsymbol{w}} \boldsymbol{x})}{\partial \hat{\boldsymbol{w}}}
\end{equation}
This scale invariance makes the key feature of the weight vector is its \emph{direction}.
When the weights are updated through stochastic gradient descent at step $t$ and learning rate $\eta$
\begin{equation}
  \boldsymbol{w}_{t+1} = \boldsymbol{w}_{t} - \eta \nabla \boldsymbol{L}_t(\boldsymbol{w}_t)
\end{equation}
As in \cite{hoffer2018norm}, one can derive that the step size of the weight direction is approximately proportional to
\begin{equation}
  \boldsymbol{\hat{w}}_{t+1} - \boldsymbol{\hat{w}}_t \propto \frac{\eta}{\norm{\boldsymbol{w}_t}^2}
\end{equation}
Based on this formulation, the ELR hypothesis can be explained as follows: when applying weight decay to layers followed by normalization,
it prevents weight norm from unlimitedly growing, which preserves the step size of weight direction, thus ``increasing the effective learning rate''

However, this phenomenon is still not fully investigated.
\citet{hoffer2018norm} propose an LR correction technique that can train DNNs to similar performance without weight decay.
However, this LR correction technique needs to mimic the effective step size \emph{from training with weight decay}.
Similar techniques have also been proposed in \citet{zhang2018three}.
These techniques act as proofs of the ELR hypothesis, but can not be used as practical training methods.
On the other hand, as the hypothesis is based on the scale invariance brought by normalizations, there are layers that do not satisfy this precondition.
For example, the final fully-connected(FC) layers that commonly used in classification tasks.
As experiments in \citet{zhang2018three}(Figure 4), there is a clear gap between whether weight decay is applied to the FC layer.
These problems indicate that the mechanisms of weight decay are still not fully understood.

\subsection{Discarding weight decay for convolution layers} \label{sec_mechanism1}

\begin{figure}[htbp]
  \centering
    \includegraphics[width=0.9\linewidth]{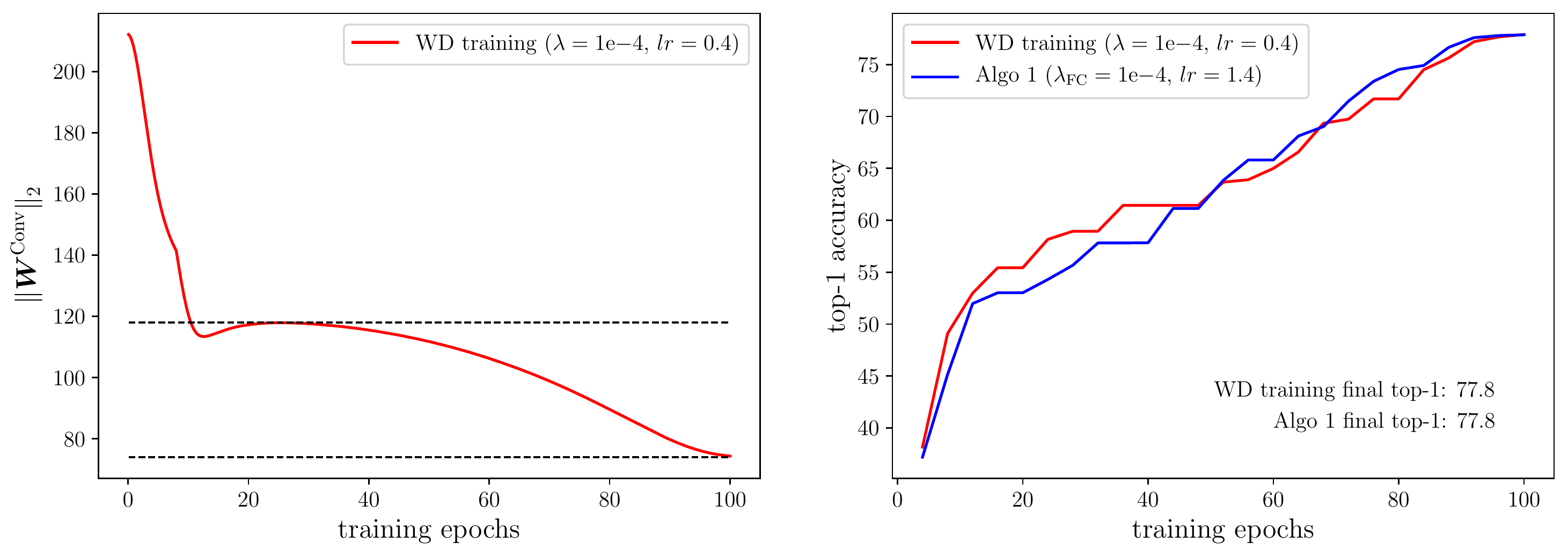}
  \caption{\textbf{Left:} $\norm{\bs{W}^{\text{Conv}}}$ of WD training \textbf{Right:} top-1 accuracy for Algo \ref{algo1} and WD training,
  both learning rates are found by gridsearch}
  \label{fig:algo1}
\end{figure}

We first consider discarding weight decay for convolution layers.
Since ELR hypothesis indicates that the main effect of weight decay on convolution layers is produced by constraining weight vector norm,
we investigate how weight vector norm changes during training.
Denoting the weights in all convolution layers as a single verctor $\bs{W}^{\text{Conv}}$, we plot $\norm{\bs{W}^{\text{Conv}}}$ in Fig \ref{fig:algo1}.
ResNet50\citet{he2016deep} is trained on ImageNet for 100 epochs with $lr=0.4$ and weght decay$\lambda=0.0001$. Other settings follow general setups in \ref{sec_exp}.

From Fig \ref{fig:algo1} left, the curve can be divided into two parts.
In the first part, $\norm{\bs{W}^{\text{Conv}}}$ decreases rapidly, which will increase the effective learning rate according to the ELR hypothesis.
However, we already adopt the learning rate warmup strategy, therefore this part of effect is duplicate and can be discarded.
In the second part, which occupies the majority of training, $\norm{\bs{W}^{\text{Conv}}}$ changes slowly in a relatively stable range.
From these observations, we propose to fix $\norm{\bs{W}^{\text{Conv}}}$ to a constant, as in Algo \ref{algo1}.
We rescale $\bs{W}^{\text{Conv}}$ after each optimization step,
which does not change outputs of the network but substantially maintains the effective learning rate.
While WD training controls the weight norm dynamically by a hyperparameter $\lambda$, Algo \ref{algo1} directly fixes the norm to a constant.
Since this constant can be any value, we choose it as the $\norm{\bs{W}^{\text{Conv}}}$ at initialization for simplicity.

\begin{algorithm}[h]
  \small
	\caption{Fixing the weight norm of convolution layers}
  \label{algo1}
  {\bf Input:} initial learning rate $lr$, total steps $T$, weight decay on final FC layer $\lambda_{\text{FC}}$, momentum $\mu$, training samples $\boldsymbol{x}$, corresponding labels $\boldsymbol{y}$ \\
	{\bf Initialization:} velocity $\boldsymbol{V}_{0} \leftarrow \boldsymbol{0}$, random initialize weight vector $\bs{W_{0}}$
\begin{algorithmic}[1]
  \For{$t$ in 0, ..., $T-1$} 
    \State $\bs{x}, \bs{y}$ $\leftarrow$ BatchSampler($t$)
    \State $\widehat{\mathcal{L}}_{t}(\bs{W}_t) \leftarrow \sum \mathcal{L}(f(x;\bs{W}_t),y)+\frac{1}{2}\lambda_{\text{FC}} \norm{\bs{W}_{t}^{\text{FC}}}^2$
    \State $\bs{V}_{t+1} \leftarrow \mu \bs{V}_t+ \nabla \widehat{\mathcal{L}}_{t}(\bs{W}_t)$
    \State $\eta_{t} \leftarrow $ GetLRScheduleMultiplier$(t)$
    \State $\bs{W}_{t+1} \leftarrow \bs{W}_{t} -lr \times \eta_t \times \bs{V}_{t+1}$
    \State $\bs{W}^{\text{Conv}}_{t+1} \leftarrow \frac{ \bs{W}^{\text{Conv}}_{t+1} }{ \norm{\bs{W}^{\text{Conv}}_{t+1}} } \norm{\bs{W}^{\text{Conv}}_{0}}$
  \EndFor
\end{algorithmic}
\end{algorithm}

Since the fixed weight norm in Algo \ref{algo1} is substantially different from WD training, their optimal learning rates are different as well.
To be fair, we grid search the $lr$ for both Algo \ref{algo1} and WD training and compare the best performance.
As shown in Fig \ref{fig:algo1} right, Algo \ref{algo1} achieves same top-1 accuracy of WD training.
These results demonstrate that for convolution layers followed by BN, weight decay can be discarded.
Note that we preserve $\lambda^{\text{FC}}$ for the final FC layer, which will be addressed in the next subsection.

\vspace{-3mm}

\subsection{Effects on final fully-connected layers} \label{sec_mechanism2}

As noted in section \ref{sec_revisit}, the final FC layer does not satisfy the precondition of the ELR hypothesis.
To investigate the effects of weight decay on this layer, we first try to make it scale-invariant.
We apply three modifications on Algo \ref{algo1}: (1) replace original FC layer with WN-FC layer; 
(2) replace $\bs{W}^{\text{Conv}}$ in line 7 of Algo \ref{algo1} with $\bs{W}^{\text{Conv+FC}}$; (3) set $\lambda^{\text{FC}}=0$.
This modified algorithm is denoted as Algo 1@WN-FC.
\footnote{For full algorithm descriptions please refer to Algo 2 in supplementary materials}
WN-FC layer is normal FC layer applied with weight normalization\citep{salimans2016weight}. Original FC and WN-FC layer are formulated as follows:

\vspace{-2mm}

\begin{align}
  \text{FC}(x; \bs{W}^{\text{FC}}) &= \trans{x} \bs{W}^{\text{FC}} \\
  \text{WN-FC}(x; \bs{W}^{\text{FC}}, g) &= \frac{\trans{x} \bs{W}^{\text{FC}}}{\norm{\bs{W}^{\text{FC}}}} \times g \label{eqa:wn-fc}
\end{align}

\begin{figure}[htbp]
  \centering
    \includegraphics[width=0.9\linewidth]{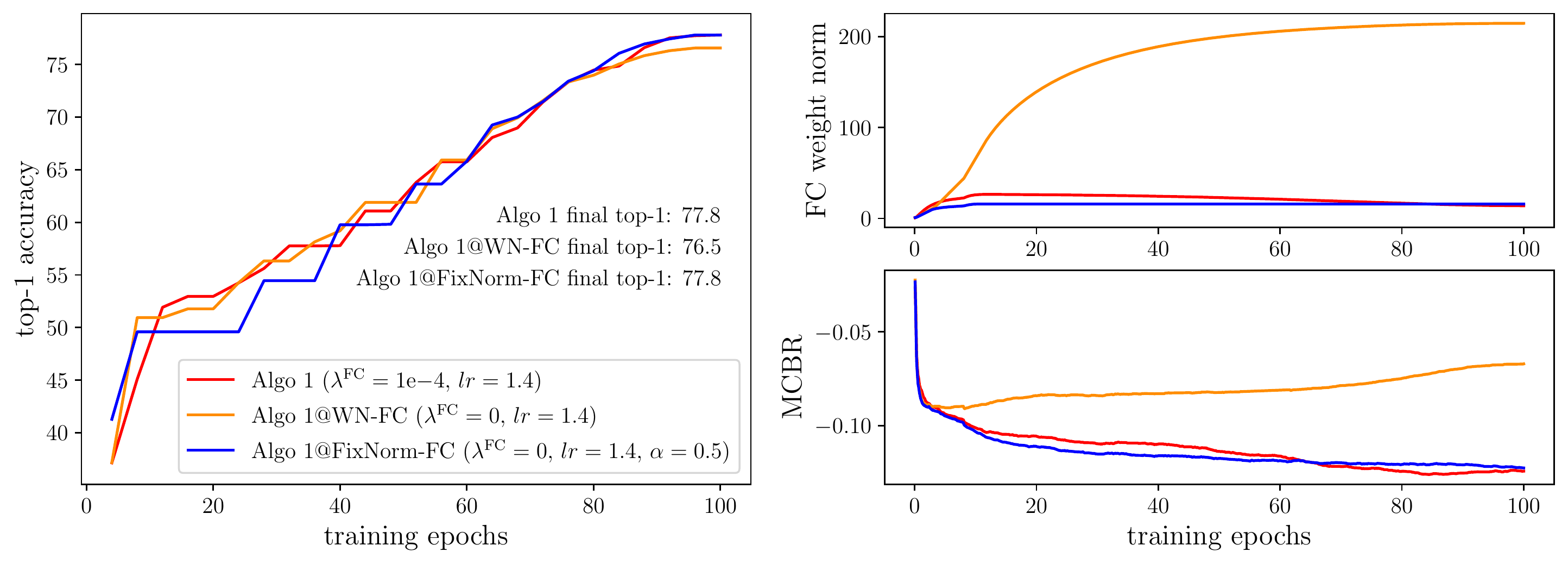}
  \caption{Training ResNet50 on ImageNet with Algo \ref{algo1}, Algo 1@WN-FC and Algo 1@FixNorm-FC. 
  \textbf{Left:} top-1 accuracy \textbf{Top right:} weight norm of the final FC layer ($\norm{\bs{W}^\text{FC}}$ for FC, $g$ for WN-FC and FixNorm-FC) 
  \textbf{Bottom right:} MCBR}
  \label{fig:algo123}
\end{figure}

We compare Algo \ref{algo1} and Algo 1@WN-FC by training ResNet50 on ImageNet.
As can be seen in Fig \ref{fig:algo123} left, there is a clear gap between Algo \ref{algo1} and Algo 1@WN-FC, 
which implies that weight decay has additional effects beyond preserving ELR on final FC layers.

Now lets combine equation \ref{eqa:wn-fc} with softmax cross-entropy loss $\mathcal{L}$, 
$s_i= \frac{\trans{x} \bs{W}_i}{\norm{\bs{W}}}g$ denotes the logits value of class $i$, 
$p_i$ denotes the probability of class $i$, $k$ for the label class and $j$ for other classes.
We have(for full derivations please refer to supplymentary materials),
\begin{align}
  \mathcal{L}(x, k) &= - \log p_k = - \log \frac{e^{s_k}}{\sum e^{s_j}} \\
  -\frac{\partial \mathcal{L}(x, k)}{\partial x} &= \frac{g}{\norm{\bs{W}}} \sum_{j \neq k} p_j(\bs{W}_k - \bs{W}_j)
\end{align}

\begin{wrapfigure}[11]{r}{0.40\textwidth}
  \centering
    \includegraphics[width=0.95\linewidth]{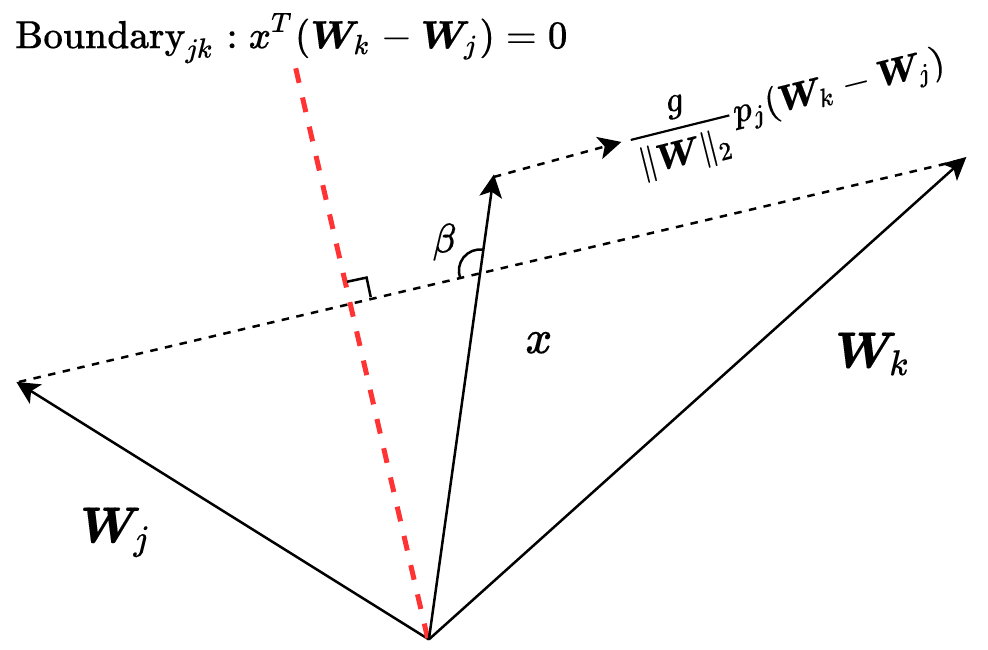}
  \caption{Ilustration of feature space}
  \label{fig:boundary}
\end{wrapfigure}

Given $\bs{W}$, the gradient is actually driving $x$ from other class center $\bs{W}_j$ towards label class center $\bs{W}_k$,
where the magnitude depends on $p_j$ and $g$.
Note that $p_j$ also depends on $g$ through the softmax function.
When $x$ is correctly classified and $g$ continuously grows, $p_j$ will rapidly shrink and weaken the gradient.
As illustrated in Fig \ref{fig:boundary}, this will leave $x$ being closer to the class boundary between
$\bs{W}_j$ and $\bs{W}_k$(larger $\cos \beta$), which is \emph{less discriminative}.
This ambiguous feature space is prone to distribution shift between training and testing, therefore may result in poor generalization.

To quantitively verify this explaination, we define \textbf{Mean Cross-Boundary Risk}(MCBR):
\begin{equation}
  \text{MCBR}(x, k, \bs{W}) = \frac{1}{\#class-1} \sum_{j\neq k} \cos(x, \bs{W}_j-\bs{W}_k)
\end{equation}
MCBR shows how much $x$ is lean to the class boundaries, ranging from -1 to 1.
The larger MCBR is, the more likely $x$ will cross the class boundaries during testing.
We compare the weight norm of FC layer($g$ for WN-FC layer) and MCBR for Algo \ref{algo1} and Algo 1@WN-FC in Fig \ref{fig:algo123} top right and bottom right.
It can be clearly observed that without constraint, $g$ continuously grows and leads to higher MCBR.
This explains why Algo 1@WN-FC generalize poorly.


Based on these analysis, we propose to constrain $g$ from exceeding a given upperbound $\alpha$, denoted as FixNorm-FC.
$\sqrt{\#class}$ normalize the upper bound across different number of classes.
\begin{equation}
  \text{FixNorm-FC}(x; \bs{W}^{FC}, g, \alpha) = \frac{\trans{x} \bs{W}^{FC}}{\norm{\bs{W}^{FC}}} \times \min (g, \alpha * \sqrt{\#class}) \label{eqa:fixnorm-fc}
\end{equation}
Note that $\bs{W}^{FC}$ and $g$ are learnable parameters, while $\alpha$ is a hyper-parameter.
We replace the WN-FC layer in Algo 1@WN-FC with the FixNorm-FC layer, denoted as Algo 1@FixNorm-FC.\footnote{For full algorithm descriptions please refer to Algo 1@FixNorm-FC in supplementary materials}
We choose $\alpha$ according to the weight norm of Algo \ref{algo1} in Fig \ref{fig:algo123} top right and leave other hyper-parameters unchanged, the results are shown in Figure \ref{fig:algo123} left.
By simply constraining the upper limit of $g$, Algo 1@FixNorm-FC maintains low MCBR and fully closes the accuracy gap.

Moreover, we find that the optimal values for $\alpha$ are different among models.
We give more comprehensive experiments in Section \ref{sec_exp}.
To summarise, for the final FC layer, the ELR hypothesis does not cover all the effects of weight decay.
We find that weight decay influences the \textbf{cross-boundary risk} by constraining the FC layer's weight norm and finally affects generalization performance.
By using the FixNorm-FC layer, Algo 1@FixNorm-FC can fully recover the accuracy of normal weight decay training.
Moreover, Algo 1@FixNorm-FC directly controls the two main mechanisms, which makes the hyperparameters more easier to tune.
We will show this in section \ref{sec_tuning} and \ref{sec_exp_tune}.

\subsection{Tuning lr and alpha for FixNorm training} \label{sec_tuning}

\begin{wrapfigure}[14]{r}{0.40\textwidth}
  \centering
    \includegraphics[width=0.95\linewidth]{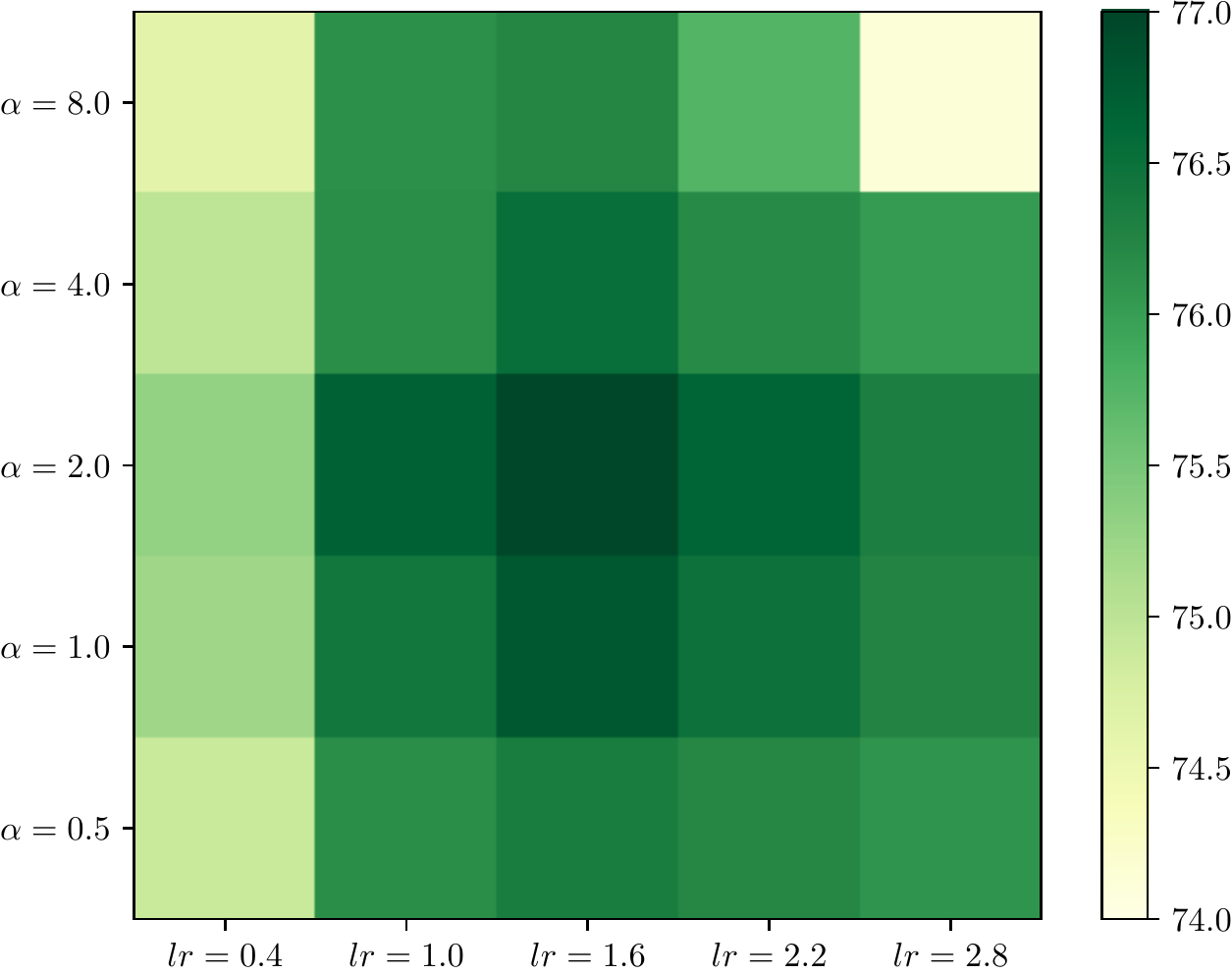}
  \caption{Top-1 accuracy for different $lr$ and $\alpha$. ResNet50 trained for 50 epochs}
  \label{fig:corr}
\end{wrapfigure}

Section \ref{sec_mechanism1} and \ref{sec_mechanism2} investigate the two main mechanisms of weight decay:
(1) for layers followed by normalizations(mainly convolution layers), affecting ELR (2) for final fully-connected layers, affecting cross-boundary risk.
Algo 1@FixNorm-FC (refered to \textbf{FixNorm} for simplicity) unifies these two mechanisms and directly controls
their effects through hyper-parameters $lr$ and $\alpha$.
While these two mechanisms both affect generalization performance, it is important to know how they are correlated,
which will determine how to tune two hyper-parameters.
To verify this, we grid search $lr$ and $\alpha$ and show the corresponding top-1 accuracy in Fig \ref{fig:corr}.
It clearly shows that the best $lr$ does not depends on the value of $\alpha$ and vice versa.
This suggests that we can tune $lr$ and $\alpha$ \emph{independently}, which will greatly reduce the cost.

Instead of using existing hyper-parameter optimization(HPO) methods, we propose a simple yet effective approach to tune $lr$ and $\alpha$.
We introduce two priors to efficiently tune $lr$.
\begin{itemize}
  \item Top-1 accuracy is approximately a convex function for $lr$
  \item The best $lr$ for shorter training is usually larger than that for longer training
\end{itemize}
The first prior is mainly an empirical finding, while the second one may be partially explained by
the correlation between generalization performance and weight distance from their initialization\citep{trainloger2017nips}:
shorter training may require larger $lr$ to travel far enough from the initialization in weight space to generalize well.
These two priors motivate us to \emph{use best lr under shorter training as an upper bound for that under longer training}.
This strategy can adaptively shrink the search range and let us locate the best $lr$ in a wide range with relatively low cost.
After the best $lr$ is found, we simply fix it and grid search for the best $\alpha$.
The overall method can be summarized in Algo \ref{algo4}(\textbf{tuned FixNorm}).

\setcounter{algorithm}{3}

\begin{algorithm}[h]
  \small
	\caption{Tuning $lr$ and $\alpha$ for FixNorm training}
  \label{algo4}
  {\bf Input:} number of $lr$ tuning rounds $N$, learning rate range $[lr_{min}, lr_{max}]$, learning rate split number $K$, training steps of each $lr$ tuning round $\bs{T}=[T_0, T_1, ..., T_{N-1}]$ where $T_i \leq T_{i+1} $, alpha candidates $[\alpha_0, \alpha_1, ..., \alpha_{m-1}]$ \\
  {\bf Output:} $\alpha_{best}$, $lr_{best}$, $acc_{best}$ \\
  {\bf Initialization:} $\alpha_{best} = \alpha_0$, $lr_{best} = \text{NULL}$, $acc_{best} = 0$
\begin{algorithmic}[1]
  \Algphase{Phase 1 - Find $lr_{best}$}
  \For{$r$ in 0, ..., $N-1$} 
    \State $\bs{LR} \leftarrow \text{UniformSplit}(lr_{min}, lr_{max}, K)$
    \State $\bs{Acc} \leftarrow \left\{ \text{FixNormTrain}(LR_k, \alpha_{best}, T_r)|\ k \in\left\{0, ..., K-1\right\} \right\}$
    \State $idx \leftarrow \arg \max \bs{Acc}$
    \State $lr_{max} \leftarrow \bs{LR}_{idx}$
    \If{$\bs{Acc}_{idx} > acc_{best} $}
      \State $acc_{best} \leftarrow \bs{Acc}_{idx}$
      \State $lr_{best} \leftarrow \bs{LR}_{idx}$
    \EndIf
  \EndFor
  \Algphase{Phase 2 - Find $\alpha_{best}$}
  \State $\bs{Acc} \leftarrow \left\{ \text{FixNormTrain}(lr_{best}, \alpha_{i}, T_{N-1})|\ i \in \left\{1, ..., m-1\right\} \right\}$
  \State $idx \leftarrow \arg \max \bs{Acc}$
  \If{$\bs{Acc}_{idx} > acc_{best} $}
    \State $acc_{best} \leftarrow \bs{Acc}_{idx}$
    \State $\alpha_{best} \leftarrow \alpha_{idx}$
  \EndIf
\end{algorithmic}
\end{algorithm}

\vspace{-1mm}

\section{Experiments} \label{sec_exp} 

\paragraph{General setups}
We perform experiments on ImageNet classification task \citep{deng2009imagenet} which contains 1.28 million training images and 50000 validation images.
Our general training settings are mainly adapted from \citet{he2019bag}, which include Nesterov Accelerated Gradient (NAG) descent\citep{nesterov1983method},
one-cycle cosine learning rate decay\citep{loshchilov2016sgdr} with linear warmup at first 4 epochs\citep{goyal2017accurate} and
label smoothing with $\epsilon=0.1$\citep{szegedy2016rethinking}.
We do not use mixup augmentation\citep{zhang2017mixup}.
All the models are trained on 16 Nvidia V100 GPUs with a total batch size of 1024. Other settings follow reference implementations of each model.
We leave experiments on MS COCO and Cityscapes in supplementary materials.

\vspace{-1mm}

\paragraph{FixNorm tuning setups}
For Algo \ref{algo4}, we set $N=2$, learning rate range $[0.2, 3.2]$, $K=5$, $\bs{T}=[0.2T_{max}, T_{max}]$ where $T_{max}$ is the max training steps, $\alpha$ candidates [0.5, 1.0, 2.0, 4.0, 8.0, 16.0].
The search contains two $lr$ tuning rounds and one $\alpha$ tuning round.
The total computational resources consumed are $K \times 0.2T_{max} + K \times T_{max} + (6-1) \times T_{max} = 11T_{max}$, which is 11 times of a single training process.

\vspace{-1mm}

\subsection{Tuned FixNorm on ResNet50-D and MobileNetV2} \label{sec_exp_tune}
To demonstrate the effectiveness of our method, we first apply Algo \ref{algo4} on two well-studied architectures: ResNet50-D\citep{he2019bag} and MobileNetV2\citep{sandler2018mobilenetv2}.
We follow \citet{he2019bag} and train 120 epochs for ResNet50-D and 150 epochs for MobileNetV2.
Their reference top-1 accuracies reported are 78.37\% and 72.04\%.
We also adopt Bayesian Optimization(BO)\citep{bo_nips2012} to search learning rate and weight decay under normal weight decay training (BO+WD) for these models.
We use the same learning rate range $[0.2, 3.2]$ for BO, and the weight decay range is set to $[0.00001, 0.0005]$.
Results are shown in Table \ref{tb:exp1}.

\vspace{-2mm}

\begin{table}[h]
  \centering
  \caption{Results for tuned FixNorm and BO+WD training}
  \label{tb:exp1}
  \resizebox{0.85\textwidth}{!}{
    \begin{tabular}{cccccccccc}
      \toprule
                    & \multicolumn{3}{c}{Tuned FixNorm} & \multicolumn{3}{c}{BO + WD} & \multicolumn{3}{c}{Reference}\\
      \cmidrule(lr){2-4} \cmidrule(lr){5-7} \cmidrule(lr){8-10}
                  & $lr$   & $\alpha$ & top-1(\%)   & $lr$   & $\lambda$ & top-1(\%) & $lr$   & $\lambda$ & top-1(\%) \\
      \midrule
      ResNet50-D   & 1.4    & 0.5 & \textbf{78.62}  & 0.53   &  8.6e-5   & 78.53     & 0.4    & 1e-4      & 78.37     \\
      MobileNetV2  & 0.5    & 16.0 & \textbf{73.20} & 0.64   &  2.2e-5   & 72.84     & 0.2    & 4e-5      & 72.04     \\
      \bottomrule
    \end{tabular}
  }
\end{table}

\vspace{-3mm}

\begin{figure}[htbp]
  \centering
  \includegraphics[width=0.9\linewidth]{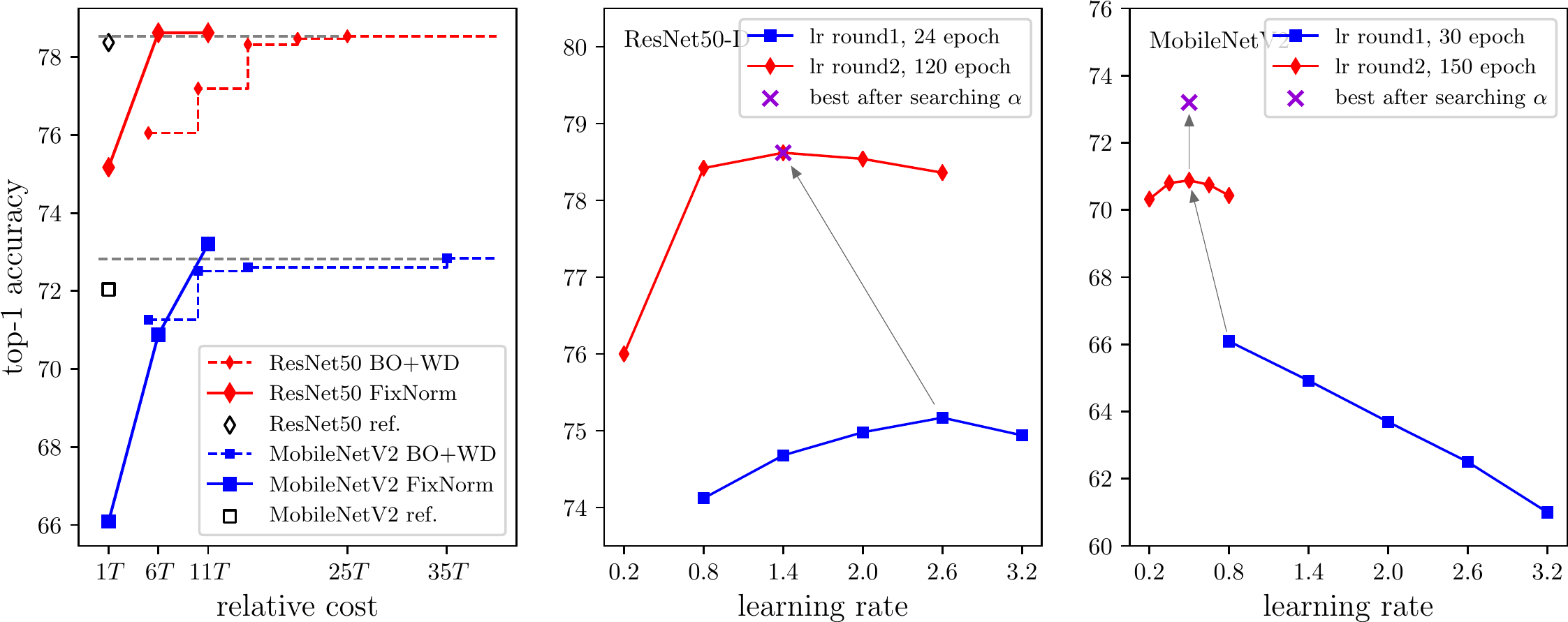}
  \caption{
    \textbf{Left:} search progress for FixNorm and BO+WD
    \textbf{Middle:} FixNorm search details for ResNet50-D 
    \textbf{Right:} FixNorm search details for MobileNetV2
  }
  \label{fig:exp1}
\end{figure}

As in Table \ref{tb:exp1}, both tuned FixNorm and BO+WD outperform reference settings by a clear margin.
Although the reference settings have already been heavily refined in \citet{he2019bag}, our method still brings substantial improvements.
We further compare tuned FixNorm and BO+WD in Fig \ref{fig:exp1} left.
Our method shows two advantages compared to BO+WD.
First, our method finds better solutions at lower cost.
The cost of our method is 11 times of normal training while BO+WD requires 25 and 35, while our final results are even better than that found by BO+WD.
Second, our method is more stable and barely needs meta-tuning.
BO itself has lots of tunable meta hyper-parameters \citep{lindauer2019towards} and it requires expert knowledge to tune them.
While we use exactly the same FixNorm tuning setups for all the experiments, including in Table \ref{tb:exp2}.
These setups are intuitive and the method performs consistently well across all the settings.

\vspace{-2mm}

\vspace{-1mm}

To better understand our method, we show more details about tuned FixNorm in Fig \ref{fig:exp1} middle and right, for ResNet50-D and MobileNetV2 respectively.
There are two $lr$ searching rounds and one $\alpha$ searching round under our setups.
In the first $lr$ round, both experiments starts with $lr$ values in [0.8, 1.4, 2.0, 2.6, 3.2] (which are the uniform split points of initial range [0.2, 3.2])
and train for $1/5$ total epochs(24 epochs and 30 epochs respectively).
Best $lr$ values are very different in this round: $lr_1=2.6$ for ResNet50-D and $lr_1=0.8$ for MobileNetV2.
Taking these values as new upper bound, [0.2, $lr_1$] are further split and corresponding $lr$ values are evaluated in the second round, for full total epochs.
The best $lr$ values are 1.4 and 0.5 in this round.
These values are fixed and then $\alpha$ is searched.
For ResNet50-D, the initial $\alpha=0.5$ is already the best, while for MobileNetV2 a better $\alpha=16.0$ is founded.
From Fig \ref{fig:exp1} one can find the patterns match two priors introduced in section \ref{sec_tuning}.

\vspace{-2mm}

\subsection{New state-of-the-arts with tuned FixNorm}

Many powerful networks have been proposed recently.
These networks usually adopt many tricks and therefore hard to tune.
To fully exploit the capabilities of these networks, we apply tuned FixNorm to further optimize them.
We also apply advanced tricks to basic models like MobileNetV2 and ResNet50-D.
These tricks are used by EfficientNet\citep{tan2019efficientnet}, including SE-layer\citep{hu2018squeeze}, swish activation\citep{ramachandran2017searching},
stochastic depth training\citep{huang2016deep} and AutoAugment preprocessing\citep{cubuk2019autoaugment}.
From Table \ref{tb:exp2}, we can find that:
\begin{itemize}
  \item Our method consistently outperforms reference settings. Strong baselines like EfficientNet can be further improved by our method, specifically +0.4\% and +0.3\% for B0 and B1.
  \item Tunning matters. When simply apply tricks to MobileNetV2 and scale to the same FLOPS with B0 and B1, tuned FixNorm achieves 77.4\% and 79.18\% top-1 accuracy,
  while default training settings only get 76.72 and 78.75. This difference can lead to unreliable conclusions when compared to EfficientNet.
  \item Best $lr$ and $\alpha$ are different among models, even for the same model with different settings. This may suggest that we should tune for each setting to fully exploit their performance.
\end{itemize}

\vspace{-2mm}

\begin{table}[h]
  \centering
  \captionof{table}{
    \textbf{Results with tuned FixNorm} models with * are applied with tricks, 
    \dag means the result is obtained using $lr$ and $\lambda$ from basic settings(MobileNetV2 150 and ResNet50-D 120)
  }
  \label{tb:exp2}
  \resizebox{0.95\textwidth}{!}{
    \begin{tabular}{ccccrrcc}
      \toprule
      Model                    &\#Epochs & Top-1      & Top-1 (ref.)  & \#Params    & \#FLOPS  & $lr$  & $\alpha$ \\
      \midrule
      MobileNetV2              & 150     & 73.20      & 72.04         &  3.5M       & 300M     & 0.5   & 16.0     \\
      MobileNetV2              & 350     & 73.97      & 73.38$^\dag$  &  3.5M       & 300M     & 0.35  & 8.0      \\
      EfficientNet-B0          & 350     & 77.72      & 77.30         &  5.3M       & 384M     & 0.5   & 4.0      \\
      EfficientNet-B1          & 350     & 79.52      & 79.20         &  7.8M       & 685M     & 0.8   & 8.0      \\
      MobileNetV2$\times$1.12* & 350     & 77.40      & 76.72$^\dag$  &  4.7M       & 386M     & 0.5   & 8.0     \\
      MobileNetV2$\times$1.54* & 350     & 79.18      & 78.75$^\dag$  &  8.0M       & 682M     & 0.65  & 4.0     \\
      ResNet50-D               & 120     & 78.62      & 78.37         &  25.6M      & 4.3G     & 1.4   & 0.5     \\
      ResNet50-D               & 350     & 79.29      & 79.04$^\dag$  &  25.6M      & 4.3G     & 1.1   & 0.5     \\
      ResNet50-D*              & 350     & 81.27      & 80.80$^\dag$  &  28.1M      & 4.3G     & 1.1   & 1.0     \\
      \bottomrule
    \end{tabular}
  }
\end{table}

\vspace{-2mm}

\section{Related works}

Due to the page limit, we only highlight the most related works in this section and leave other works in supplementary materials.

\paragraph{Understanding weight decay} Recently, a series of works \citep{van2017l2, zhang2018three, hoffer2018norm} propose that when combined with normalizations,
the main effect of weight decay is increasing ELR, which is contrary to the previous understanding and motivates new perspectives.
\citet{van2017l2} first introduces the ELR hypothesis and provides derivations for different optimizers,
while both \citet{hoffer2018norm,zhang2018three} give additional evidence supporting the hypothesis.
\citet{hoffer2018norm} also proposes norm-bounded Weight Normalization, which fixes the norm of each convolution layer separately.
By doing this, their method fixes the ELR of each layer, which \emph{highly depends on the initialization of each layer}.
Differently, we fix the norm of all convolution layers as a whole and maintains the \emph{global} ELR,
which is more robust and demonstrates SOTA performance on large scale experiments.
Layer-wise ELR controlling is an interesting problem and may lead to new perspectives for weight initialization techniques.
Similar to \citet{hoffer2018norm}, \citet{xiang2019understanding} also proposes modifications to Weight Normalization based on ELR hypothesis.
They identify the problems when using weight decay with Weight Normalization, and propose $\epsilon-$shifted $L_2$ regularizer
to constrain weight norm to $\epsilon$ with coefficient $\lambda$.
Beyond ELR hypothesis, \citet{li2019exponential} derives a closed-form between learning rate, weight decay and momentum,
and proposes an exponentially increasing learning rate schedule.
Their work mainly discusses the linkage of three hyper-parameters, while our work focuses on the underlying mechanisms of weight decay.
Except for the ELR hypothesis, \citet{loshchilov2017decoupled} identifies problems when applying weight decay to Adam optimizer,
which improves generalization performance and decouples it from learning rate.
All these works bring interesting perspectives for understanding weight decay, yet our work has distinct differences and contributions.
First, our work investigates the effect on final FC layers and find a new mechanism that complements the understanding of weight decay
on generalization performance, which is mostly ignored by previous works.
Second, our method including FixNorm and the tuning method are both concise and effective, and demonstrate SOTA performance on large scale datasets.


\section{Conclusion}
In this paper, we find a new mechanism of weight decay on final FC layers,
which affects generalization performance by controlling \emph{cross-boundary risk}.
This new mechanism complements the ELR hypothesis and gives a better understanding of weight decay.
We propose a new training method called \textbf{FixNorm}, which discards weight decay and directly controls the two mechanisms.
We also propose an effective, efficient and robust method to tune hyperparameters of FixNorm, which can consistently find near-optimal solutions in a few trials.
Experiments on large scale datasets demonstrate our methods, and a series of SOTA baselines are established for fair comparisons.
We believe this work brings new perspectives and may motivate interesting ideas like controlling layer-wise ELR and automatically adjusting cross-boundary risk.

\bibliography{iclr2021_conference}
\bibliographystyle{iclr2021_conference}

\appendix
\section{Appendix}

\subsection{Algorithms}

In section 2.3, we apply three modifications to Algo 1: (1) replace original FC layer with WN-FC layer; 
(2) replace $\bs{W}^{\text{Conv}}$ in line 7 of Algo \ref{algo1} with $\bs{W}^{\text{Conv+FC}}$; (3) set $\lambda^{\text{FC}}=0$.
This modified algorithm is denoted as Algo 1@WN-FC. Here we show the full Algo 1@WN-FC as follows:

\begin{algorithm}[H]
  \small
	\caption{@WN-FC}
  \label{algo1}
  {\bf Input:} initial learning rate $lr$, total steps $T$, momentum $\mu$, training samples $\boldsymbol{x}$, corresponding labels $\boldsymbol{y}$ \\
  {\bf Replace:} replace original FC layer with WN-FC layer \\
  {\bf Initialization:} velocity $\boldsymbol{V}_{0} \leftarrow \boldsymbol{0}$, random initialize weight vector $\bs{W_{0}}$ 
\begin{algorithmic}[1]
  \For{$t$ in 0, ..., $T-1$} 
    \State $\bs{x}, \bs{y}$ $\leftarrow$ BatchSampler($t$)
    \State $\widehat{\mathcal{L}}_{t}(\bs{W}_t) \leftarrow \sum \mathcal{L}(f(x;\bs{W}_t),y)$
    \State $\bs{V}_{t+1} \leftarrow \mu \bs{V}_t+ \nabla \widehat{\mathcal{L}}_{t}(\bs{W}_t)$
    \State $\eta_{t} \leftarrow $ GetLRScheduleMultiplier$(t)$
    \State $\bs{W}_{t+1} \leftarrow \bs{W}_{t} -lr \times \eta_t \times \bs{V}_{t+1}$
    \State $\bs{W}^{\text{Conv+FC}}_{t+1} \leftarrow \frac{ \bs{W}^{\text{Conv+FC}}_{t+1} }{ \norm{\bs{W}^{\text{Conv+FC}}_{t+1}} } \norm{\bs{W}^{\text{Conv+FC}}_{0}}$
  \EndFor
\end{algorithmic}
\end{algorithm}

The Algo 1@FixNorm-FC is similar to Algo 1@WN-FC, the only different is that we use FixNorm-FC layer instead of WN-FC layer. The full algorithm is shown as follows:

\begin{algorithm}[H]
  \small
	\caption{@FixNorm-FC}
  \label{algo1}
  {\bf Input:} initial learning rate $lr$, total steps $T$, momentum $\mu$, training samples $\boldsymbol{x}$, corresponding labels $\boldsymbol{y}$ \\
  {\bf Replace:} replace original FC layer with FixNorm-FC layer \\
  {\bf Initialization:} velocity $\boldsymbol{V}_{0} \leftarrow \boldsymbol{0}$, random initialize weight vector $\bs{W_{0}}$ 
\begin{algorithmic}[1]
  \For{$t$ in 0, ..., $T-1$} 
    \State $\bs{x}, \bs{y}$ $\leftarrow$ BatchSampler($t$)
    \State $\widehat{\mathcal{L}}_{t}(\bs{W}_t) \leftarrow \sum \mathcal{L}(f(x;\bs{W}_t),y)$
    \State $\bs{V}_{t+1} \leftarrow \mu \bs{V}_t+ \nabla \widehat{\mathcal{L}}_{t}(\bs{W}_t)$
    \State $\eta_{t} \leftarrow $ GetLRScheduleMultiplier$(t)$
    \State $\bs{W}_{t+1} \leftarrow \bs{W}_{t} -lr \times \eta_t \times \bs{V}_{t+1}$
    \State $\bs{W}^{\text{Conv+FC}}_{t+1} \leftarrow \frac{ \bs{W}^{\text{Conv+FC}}_{t+1} }{ \norm{\bs{W}^{\text{Conv+FC}}_{t+1}} } \norm{\bs{W}^{\text{Conv+FC}}_{0}}$
  \EndFor
\end{algorithmic}
\end{algorithm}

\subsection{Derivations}

The complete derivations of equation 8 are as follows.
$\mathcal{L}$ denotes the softmax cross-entropy loss,
$s_i= \frac{\trans{x} \bs{W}_i}{\norm{\bs{W}}}g$ denotes the logits value of class $i$, 
$p_i$ denotes the probability of class $i$, $k$ for the label class and $j$ for other classes.
We have,
\begin{align}
  \mathcal{L}(x, k) &= - \log p_k = - \log \frac{e^{s_k}}{\sum e^{s_i}} \\
\end{align}

We first derive $\frac{\partial p_k}{\partial s_k}$:

\begin{align}
  \frac{\partial p_k}{\partial s_k} &= \frac{
    e^{s_k} \sum e^{s_j} - e^{s_k} e^{s_k}
  }{
    \left( \sum e^{s_j} \right)^2
  } \\
  & = \frac{e^{s_k}}{\sum e^{s_j}} - \left( \frac{e^{s_k}}{\sum e^{s_j}} \right)^2 \\
  & = p_k - {p_k}^2 \\
  & = p_k (1 - p_k)
\end{align}

also for $j \neq k$, we have,

\begin{align}
  \frac{\partial p_k}{\partial s_j} &= \frac{
    - e^{s_k} e^{s_j}
  }{
    \left( \sum e^{s_i} \right)^2
  } \\
  & = - p_k p_j 
\end{align}

combine them with $\frac{\partial s_i}{\partial x} = \frac{\bs{W}_i}{\norm{\bs{W}}}g$, we have,
\begin{align}
  \frac{\partial p_k}{\partial x} &= \frac{\partial p_k}{\partial s_k} \frac{\partial s_k}{\partial x} + 
    \sum_{j \neq k} \frac{\partial p_k}{\partial s_j} \frac{\partial s_j}{\partial x} \\
    &= p_k (1 - p_k) \frac{\bs{W}_k}{\norm{\bs{W}}}g + \sum_{j \neq k} - p_k p_j \frac{\bs{W}_j}{\norm{\bs{W}}}g \\
    &= \frac{p_k g}{\norm{\bs{W}}} \left( (1-p_k) \bs{W}_k + \sum_{j \neq k} -p_j \bs{W}_j \right) \\
    &= \frac{p_k g}{\norm{\bs{W}}} \left( \bs{W}_k + \sum -p_i \bs{W}_i \right) \\
    &= \frac{p_k g}{\norm{\bs{W}}} \left( \sum p_i (\bs{W}_k - \bs{W}_i) \right) \\
    &= \frac{p_k g}{\norm{\bs{W}}} \left( \sum_{j \neq k} p_j (\bs{W}_k - \bs{W}_j) \right) \\
\end{align}

and finally,

\begin{align}
  -\frac{\partial \mathcal{L}(x, k)}{\partial x} &= -\frac{\partial \mathcal{L}(x, k)}{\partial p_k} \frac{\partial p_k}{\partial x} \\
   &= \frac{1}{p_k} \frac{p_k g}{\norm{\bs{W}}} \sum_{j \neq k} p_j(\bs{W}_k - \bs{W}_j) \\
   &= \frac{g}{\norm{\bs{W}}} \sum_{j \neq k} p_j(\bs{W}_k - \bs{W}_j)
\end{align}

\subsection{More Details}

\paragraph*{Parameters other than convolution and FC weights} For modern CNNs like ResNet or MobileNet, the majority of parameters come from 
weights of convolution and FC layers. Other parameters are mainly biases and $\gamma$ and $\beta$ of BN layers.
As in \cite{he2019bag}, the no-bias-decay strategy is applied to avoid overfitting, which does not use weight decay on these parameters.
We empirically find that this strategy does not harm performance, so we adopt this strategy in our FixNorm method,
which means we do not fix the norm of biases and $\gamma$ and $\beta$ parameters.
Experiments in Table 2 also include architectures with SE-blocks, which have FC layers that are not followed by normalizations.
Since these layers are not directly followed by softmax cross-entropy loss, we find that they do not suffer from the problem identified in section 2.3.
So we simply replace these layers with WN-FC layers and add the weights into the norm-fixing process.
In summary, our FixNorm method considers weights of convolution layers, final FC-layers, and FC layers of SE-blocks.
Other parameters like biases and $\gamma$ and $\beta$ of BN layers are excluded from the norm fixing process.

\subsection{More results on segmentation and object detection}

\subsubsection{Extending FixNorm-FC for pixel-wise classification}

The FixNorm-FC is proposed to replace the original final FC layer in classification tasks.
There are other forms of classification tasks that do not use FC layers, such as segmentation and object detection.
For segmentation, the models are usually fully convolutional and the last convolution layer is used for pixel-wise classification.
This also applies to Region Proposal Networks\cite{ren2015faster} used in object detection or methods that produce dense detections like RetinaNet\cite{lin2017focal}.
These tasks still share the nature of the classification task, therefore the cross-boundary risk still needs to be controlled.
Our FixNorm-FC layer can be easily extended to these tasks because the final convolution layer can be viewed as a normal FC layer
that shares weight across spatial positions. Denote the weight of the final convolution layer as $\bs{W}^{\text{Conv}}$ with shape $[c_{out}, c_{in}, k_h, k_w]$,
we define,

\begin{equation}
  \text{FixNorm-Conv}(x; \bs{W}^{\text{Conv}}, g, \alpha) = \frac{\text{Conv}(x, \bs{W}^{\text{Conv}})}{\norm{\bs{W}^{\text{Conv}}}} \times \min (g, \alpha * \sqrt{c_{out}}) \label{eqa:fixnorm-conv}
\end{equation}

This layer is a straight forward extension of FixNorm-FC layer, which will be used in experiments on segmentation and detection later.

\subsubsection{Experiments on Cityscapes}

\paragraph{Setups} The Cityscapes dataset\cite{Cordts2016Cityscapes} is a task for urban scene understanding.
We follow the basic training settings in \cite{yuan2018ocnet}.
We use 2975 images for training and 500 images for validation.
The initial learning rate is set as 0.01 and weight decay as 0.0005.
The original image size is 1024$\times$2048 and we use crop size of 769$\times$769.
All the models are trained on 4 Nvidia V100 GPUs for 40000 iterations with a total batch size of 8.
The poly learning rate policy is used. We use the ResNet-101 + Base-OC\cite{yuan2018ocnet} as the baseline model.

\paragraph{Modifications}
We replace the last convolution layer with FixNorm-Conv when trained with our FixNorm method.

\paragraph{FixNorm tuning setups} The main settings are the same with that on ImageNet, such as $N=2$, $K=5$, $T=[0.2T_{max}, T_{max}]$,
$\alpha$ candidates as [0.5, 1.0, 2.0, 4.0, 8.0, 16.0].
The only difference is that we adapt the learning rate range to [0.005, 0.1].
The reason is that the models are finetuned on a pre-trained model from ImageNet, therefore the default learning rate is smaller.

The results are shown in table \ref{tb:seg}. Tuned FixNorm clearly outperforms the baseline and the improvements are larger when the
cosine learning rate is applied.

\begin{table}[h]
  \centering
  \captionof{table}{
    \textbf{Results with tuned FixNorm on Cityscapes}
  }
  \label{tb:seg}
  \resizebox{0.95\textwidth}{!}{
    \begin{tabular}{cccl}
      \toprule
      Model                      &\#Iters & Val. mIoU(\%)  &   hyperparameters  \\
      \midrule
      baseline                   & 40000  & 78.7           &   lr=0.01, wd=0.0005     \\
      baseline w/ cosine lr      & 40000  & 78.3           &   lr=0.01, wd=0.0005     \\
      tuned FixNorm              & 40000  & 79.4           &   lr=0.0335, $\alpha=1.0$  \\
      tuned FixNorm w/ cosine lr & 40000  & 79.7           &   lr=0.043, $\alpha=1.0$ \\
      \bottomrule
    \end{tabular}
  }
\end{table}

\subsubsection{Experiments on MS COCO}

\paragraph{Setups} To verify our tuned FixNorm method on object detection task, we train RetinaNet\cite{lin2017focal} on MS COCO\cite{lin2014microsoft}.
We following common practice and use the COCO trainval35k split, and report results on the minival split.
We use the ResNet50-FPN backbone, while the base ResNet50 model is pre-trained on ImageNet.
The RetinaNet is trained with stochastic gradient descent(SGD) on 8 Nvidia V100 GPUs with a total batch size of 16.
The models are trained for 90k iterations with default learning rate 0.01, which is then divided by 10 at 60k and 80k iterations.
The default weight decay is 0.0001. The $\alpha_{focal}$ is set to 0.25 and the $\gamma_{focal}$ is set to 2.0.
The standard smooth $L_1$ loss is used for box regression.
We use horizontal image flipping as the only data augmentation, and the image scale is set to 800 pixels.

\paragraph{Modifications}
To make the RetinaNet compatible with our method, we add Weight Normalization layers to
all the convolution layers that are not followed by normalizations (include layers in FPN and classification subnet and bounding-box prediction subnet),
for all the models.
We also replace the last convolution layer of the classification subnet with FixNorm-Conv when trained with our FixNorm method.
The last convolution layer of the bounding-box prediction subnet is used for regression task, which does not suffer from the problem identified
in section 2.3, so we do not replace it with FixNorm-Conv.

\paragraph{FixNorm tuning setups} The main settings are the same with that on ImageNet, such as $N=2$, $K=5$, $T=[0.2T_{max}, T_{max}]$,
$\alpha$ candidates as [0.5, 1.0, 2.0, 4.0, 8.0, 16.0].
As the models are finetuned on a pre-trained model, we use the same learning rate range of [0.005, 0.1] as in segmentation experiments.

The results are shown in table \ref{tb:det}.
Tuned FixNorm clearly outperforms the baseline and the improvements are larger when the
cosine learning rate is applied.

\begin{table}[H]
  \centering
  \captionof{table}{
    \textbf{Results with tuned FixNorm on MS COCO}
  }
  \label{tb:det}
  \resizebox{0.95\textwidth}{!}{
    \begin{tabular}{cccl}
      \toprule
      Model                      &\#Iters & Val. AP(\%)  &   hyperparameters  \\
      \midrule
      baseline                   & 90000  & 36.5         &   lr=0.01, wd=0.0001      \\
      baseline w/ cosine lr      & 90000  & 36.2         &   lr=0.01, wd=0.0001      \\
      tuned FixNorm              & 90000  & 36.9         &   lr=0.0145, $\alpha=0.5$  \\
      tuned FixNorm w/ cosine lr & 90000  & 37.1         &   lr=0.0145, $\alpha=0.5$  \\
      \bottomrule
    \end{tabular}
  }
\end{table}

\subsection{Additional related works}

\textbf{Hyperparameter Optimization (HPO).} Hyperparameter Optimization is an important topic for effectively training DNNs.
One straight forward method is grid search, which is only affordable for a very limited number of hyperparameters since the combinations grow exponentially.
Random search\citep{bull2011convergence} is a popular alternative that select hyperparameter combinations randomly,
which can be more efficient when the resource is constrained.
Bayesian Optimization(BO) \citep{brochu2010tutorial} further improves efficiency by using a model that is built on historical information to guide the selection.
Hyperband\citep{li2017hyperband} allocates different budgets to random configurations and rejects bad ones according to the performance obtained under low budgets.
BOHB \citep{falkner2018bohb} combines BO with Hyperband to select more promising configurations.
Both Hyperband and BOHB highly relies on the assumption that performance under different budgets is consistent.
However, this assumption is not always true and these methods may suffer from the low rank-correlation of performance under different budgets\citep{ying2019bench}.
While these methods are universal black-box optimization methods, our tuning method leverages more priors of hyperparameters.
Our method suggests that with a better understanding of the underlying mechanisms, we can develop a method that is both effective and efficient.

\end{document}